\documentclass{article}
\usepackage{iclr2026_conference,times}

\usepackage{amsmath,amsfonts,bm}

\def\eqref#1{equation~\ref{#1}}

\def\1{\bm{1}}

\DeclareMathAlphabet{\mathsfit}{\encodingdefault}{\sfdefault}{m}{sl}
\SetMathAlphabet{\mathsfit}{bold}{\encodingdefault}{\sfdefault}{bx}{n}

\usepackage{hyperref}
\usepackage{url}
\usepackage{graphicx}
\usepackage{booktabs}
\usepackage{multirow}
\usepackage{makecell}
\usepackage{placeins}
\usepackage[most]{tcolorbox}
\usepackage{caption}

\title{
Benchmarking LLM Summaries of Multimodal Clinical Time Series for Remote Monitoring
}
\author{
\textbf{Aditya Shukla}$^{1*}$,
\textbf{Yining Yuan}$^{1*}$,
\textbf{Ben Tamo}$^{1}$,
\textbf{Yifei Wang}$^{1}$, \\
\textbf{Micky Nnamdi}$^{1}$,
\textbf{Shaun Tan}$^{1}$,
\textbf{Jieru Li}$^{1}$,
\textbf{Benoit Marteau}$^{1}$, \\
\textbf{Brad Willingham}$^{2}$,
\textbf{May D. Wang}$^{1}$ \\[0.5em]
$^{1}$Georgia Institute of Technology \\
$^{2}$Shepherd Center \\[0.5em]
\texttt{\{ashukla301, yyuan394, jtamo3, ywang4343, mnnamdi3,} \\
\texttt{stan99, jli3545, benoitmarteau, maywang\}@gatech.edu}\\
\texttt{brad.willingham@shepherd.org}
}
\track{Research}
 \iclrfinalcopy
\begin{document}
\raggedbottom
\maketitle
\begin{center}
\small{$^{*}$Equal contribution.}
\end{center}
\begin{abstract}
Large language models (LLMs) can generate fluent clinical summaries of remote therapeutic monitoring time series, yet it remains unclear whether these narratives faithfully capture clinically significant events such as sustained abnormalities. Existing evaluation metrics emphasize semantic similarity and linguistic quality, leaving event-level correctness largely unmeasured. We introduce an event-based evaluation framework for multimodal time-series summarization using the technology-integrated health management (TIHM)-1.5 dementia monitoring data. Clinically grounded daily events are derived via rule-based abnormal thresholds and temporal persistence, and model-generated summaries are aligned to these structured facts. Our protocol measures abnormality recall, duration recall, measurement coverage,  and hallucinated event mentions. 
Benchmarking zero-shot, statistical prompting, and vision-based pipeline using rendered time-series visualizations reveals a striking decoupling: models with high conventional scores often exhibit near-zero abnormality recall, while the vision-based approach achieves the strongest event alignment (45.7\% abnormality recall; 100\% duration recall).
These results highlight the need for event-aware evaluation to ensure reliable clinical time-series summarization.
\end{abstract}

\section{Introduction}
Home-based remote therapeutic monitoring (RTM) is increasingly used to support people living with dementia, combining physiological devices (e.g., blood-pressure cuffs, scales) with ambient sensors, such as bed, motion, and door sensors. Large deployments such as technology-integrated health management (TIHM)-1.5 collect continuous multimodal time series that can reveal early deterioration and support proactive care~\citep{palermo2023tihm}. 
However, in current clinical workflows, practitioners must manually synthesize these data across disparate plots and alert logs, mentally integrating trends across multiple modalities and temporal scales, a process that is both cognitively demanding and prone to oversight.

Language models have introduced powerful new primitives for time-series analysis. Foundational models such as Lag-Llama, Chronos, and TimesFM have set new benchmarks for zero-shot numerical forecasting by capturing deep temporal dependencies \citep{rasul2023lag, ansari2024chronos, das2024decoder}. Simultaneously, general-purpose Large Language Models (LLMs) like GPT-4o have demonstrated sophisticated reasoning and summarization capabilities \citep{achiam2023gpt}. Despite these gains, a critical validation gap remains: while these models excel at capturing statistical patterns or following linguistic instructions, their ability to generate clinically faithful prose, grounded in raw sensor evidence, remains unproven.

Current evaluation paradigms for LLM summarization typically rely on semantic overlap (e.g., ROUGE, BERTScore, Moverscore) or Natural Language Inference (NLI)-based consistency metrics such as SummaC or AlignScore \citep{lin2004rouge, zhang2019bertscore, zhao2019moverscore, laban-etal-2022-summac, zha-etal-2023-alignscore}. However, these metrics prioritize linguistic fluency and topical similarity over the precise, threshold-based factual accuracy required in medical monitoring. In the context of dementia care, the omission of a subtle "sustained deviation" (e.g., a patient remaining in bed for an abnormal duration) is a catastrophic failure that standard NLP metrics are poorly equipped to detect.

To bridge this gap, we introduce an event-based evaluation framework designed to measure clinical faithfulness in multimodal time-series narratives. Leveraging the TIHM-1.5 dataset, we construct a rule-grounded evaluation system that extracts structured clinical events, such as abnormal vital sign means and sustained behavioral deviations, to serve as an immutable ground truth. We then assess LLM performance through the lens of event-level factual recall, duration accuracy, and modality coverage, allowing for a rigorous decomposition of omissions versus hallucinations.


\textbf{Main contributions}. This work (1) introduces an event-based evaluation framework for multimodal clinical time-series summarization, measuring abnormality recall, duration faithfulness, and modality coverage, (2) demonstrates a striking decoupling between conventional summarization metrics and event-level clinical correctness on TIHM-1.5 dementia monitoring data, and (3) benchmarks prompting- and visualization-grounded multimodal pipelines, showing that summaries based on rendered time-series plots achieve stronger alignment with underlying physiological events.



\begin{figure}
    \centering
    \includegraphics[width=1\linewidth]{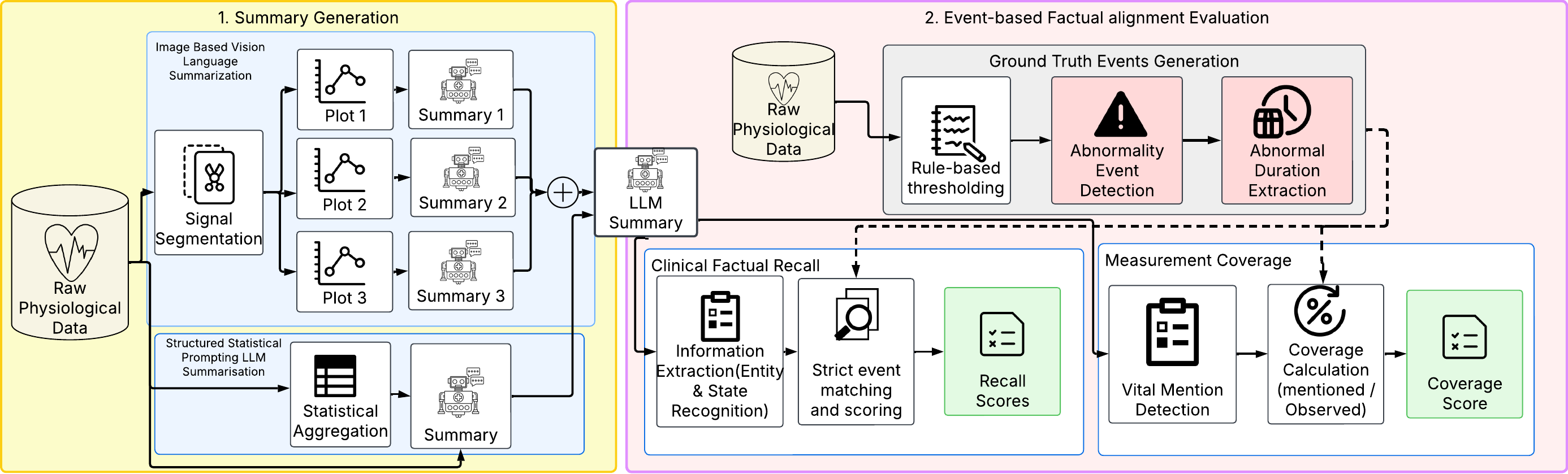}
    \caption{\textbf{Overview of the proposed event-based evaluation framework for clinical time-series summarization.} 
\textbf{Left:} Summary generation pipelines, including (1) image-based vision--language summarization via signal segmentation and plot rendering, and (2) structured statistical prompting with aggregated numerical features. Both pipelines generate daily LLM summaries from raw physiological data. 
\textbf{Right:} Event-based factual alignment evaluation. Ground-truth clinical events are derived through rule-based thresholding to detect abnormalities and sustained durations. Generated summaries are evaluated using strict event matching (clinical factual recall) and measurement coverage analysis, producing recall and coverage scores that quantify event-level correctness beyond reference-based semantic similarity metrics.}

    \label{fig:overview}
\end{figure}

\section{Methodology}

\subsection{Dataset and Problem Setup}

TIHM-1.5~\citep{palermo2023tihm} is a public remote monitoring dataset covering 56 people living with dementia across 2{,}803~patient-days, recording physiology (heart rate, blood pressure, temperature), room-level activity, and sleep staging with cardiorespiratory signals. We define each patient-day as the unit of analysis:
\begin{equation}
    \mathcal{D}_{i,t} = \{x^{(m)}_{i,t}\}_{m=1}^{M},
\end{equation}
where $x^{(m)}_{i,t}$ denotes timestamped observations from modality $m$. A stratified test set of $N{=}100$ patient-days is held out for evaluation; remaining days are reserved for prompt development. Given $\mathcal{D}_{i,t}$, the goal is to generate a clinically grounded narrative $y_{i,t}$; pipelines differ only in how the time series is represented to the model.

\subsection{Summarization Pipelines}
\paragraph{Raw signal prompting.} The LLM receives a direct textual serialization of $\mathcal{D}_{i,t}$ and must perform all numerical reasoning internally.

\paragraph{Statistical conditioning.} Because LLMs struggle with long
floating-point sequences, we also condition on a compact summary vector
$S_{i,t} = f(\mathcal{D}_{i,t})$, where $f(\cdot)$ extracts per-modality
descriptive statistics (mean, min, max, std) and discrete abnormality
indicators.

\paragraph{Visualization grounding.} Each patient-day is rendered into
clinical-style plots $I_{i,t} = r(\mathcal{D}_{i,t})$, annotated with
thresholds and sleep-stage context. A vision-language model generates
$y_{i,t}$ from $I_{i,t}$, reporting only visually supported events.

\noindent Across all settings, the output is a structured daily narrative
$y_{i,t}$, conditioned on $\mathcal{D}_{i,t}$, $S_{i,t}$, or $I_{i,t}$
respectively.

\subsection{Ground-Truth Fact Extraction}

We derive clinical event facts from established clinical standard vital-sign reference ranges without manual annotation. For each patient-day, we instantiate an \emph{abnormality fact} when any vital falls outside predefined bounds (e.g., HR ${<}50$ or ${>}90$\,bpm) and a \emph{duration fact} when out-of-range values persist for ${\geq}\Delta{=}30$\,minutes. Each patient-day yields a structured fact set
\begin{equation}
    F_{i,t} = \{f_k\}_{k=1}^{K},
\end{equation}
encoding (\textit{vital}, \textit{type}, \textit{direction}, \textit{value/duration}), together with a record of modality availability.

\subsection{Event-Based Evaluation}

Generated summaries are aligned to $F_{i,t}$ using a conservative mention
policy requiring co-occurrence of the correct vital name and an explicit
abnormality indicator (e.g., ``high blood pressure,'' ``outside normal
range''). We report two complementary metrics:
\begin{equation}
\mathrm{Recall}(y_{i,t}) =
\frac{|\{f \in F_{i,t} : f \text{ is mentioned in } y_{i,t}\}|}
     {|F_{i,t}|},
\end{equation}
\begin{equation}
\mathrm{Coverage}(y_{i,t}) =
\frac{|\{m : x^{(m)}_{i,t} \text{ observed and mentioned}\}|}
     {|\{m : x^{(m)}_{i,t} \text{ observed}\}|}.
\end{equation}
Recall quantifies omission of clinically significant events; coverage
measures whether all observed modalities are acknowledged. Claims matching
no fact in $F_{i,t}$ are flagged as hallucinations.

\section{Experiments and Results}
\subsection{Experimental Setup}

We evaluate generated summaries along three complementary dimensions: event-level factual alignment, semantic faithfulness, and perceived clinical clarity.

\textbf{Event-Level Factual Alignment:} We decompose clinical faithfulness into three granular metrics: (i) \textit{Abnormality Recall}, measuring the retrieval of point-wise threshold violations; (ii) \textit{Duration Recall}, measuring the identification of sustained deviations (persisting $>30$ mins); and (iii) \textit{Measurement Coverage}, quantifying the model's acknowledgment of available data streams versus hallucinated omissions.

\textbf{Semantic faithfulness.} 
We further evaluate reference-free consistency using AlignScore~\citep{zha-etal-2023-alignscore} and SummaC~\citep{laban-etal-2022-summac}, treating the structured tabular cues as the source document and the generated summary as the hypothesis.

\textbf{LLM-as-judge clarity.} 
We utilize GPT-4o-mini as a proxy evaluator, scoring summaries on a 1--5 Likert scale for coherence, readability, and professional tone.

\subsection{Results}
\begin{table*}[t]
\centering
\caption{Event-level clinical correctness diverges from conventional NLP metrics. Statistical conditioning improves grounding, and the visualization-grounded pipeline achieves the strongest event alignment despite lower NLP scores.}
\label{tab:main_results}
\resizebox{\linewidth}{!}{
\begin{tabular}{l|l |ccc |ccc}
\toprule

\multirow{2}{*}{\textbf{Method}} 
& \multirow{2}{*}{\textbf{Backbone}} 
& \multicolumn{3}{c}{\textbf{Clinical Event Metrics - Ours} (\%)} 
& \multicolumn{3}{|c}{\textbf{Standard NLP Metrics} ($\uparrow$)} \\
\cmidrule(lr){3-5} \cmidrule(lr){6-8}
& 
& Abnormality 
& Duration
& Coverage
& SummaC 
& Align
& Clarity \\
\midrule

\multirow{4}{*}{Zero-shot}
& Llama-3.2-3B        
& 0.00
& 0.00
& 46.44
& 0.59
& 0.32
& 3.27 \\

& Llama-3-8B        
& 0.00
& 0.00
& 41.40
& 0.57
& 0.31
& \textbf{4.00} \\

& Gemma-3-12B       
& 0.00
& 0.00 
& 31.50 
& 0.28
& 0.39
& 1.70 \\

& Gemma-3-27B       
& 2.10 
& 0.00 
& 15.60 
& 0.55 
& \textbf{0.44} 
& 1.44 \\

\midrule

\multirow{4}{*}{Stat Based}
& Llama-3.2-3B        
& 18.80
& 0.00
& 99.32
& 0.44
& 0.23
& 3.38 \\

& Llama-3-8B        
& 33.30
& 40.00
& 94.20
& 0.64
& 0.28
& \textbf{4.00} \\

& Gemma-3-12B       
& 27.10 
& 20.00 
& 56.61
& \textbf{0.65}
& 0.39 
& 1.87 \\

& Gemma-3-27B       
& 36.70
& 60.00
& 15.60 
& 0.55
& 0.39
& 1.79 \\

\midrule

\multirow{1}{*}{Image-Based}
& Gemini-2.5-Pro 
& \textbf{45.70} 
& \textbf{100.00} 
& \textbf{100.00} 
& 0.38 
& 0.16 
& 2.71 \\
\bottomrule
\end{tabular}
}
\end{table*}



Table~\ref{tab:main_results} reveals a striking decoupling between conventional summarization metrics and event-level clinical correctness. Zero-shot text-only models achieve moderate semantic faithfulness and high clarity scores, yet fail almost entirely to report clinically significant events, with near-zero abnormality recall (0.0--2.1\%) and no detected duration abnormalities. In contrast, statistical conditioning substantially improves grounding: Llama-3-8B rises from 0.0\% to 33.3\% abnormality recall and reaches 94.2\% measurement coverage, suggesting that quantitative signal extraction is a major bottleneck for text-based LLMs. The visualization-grounded Gemini-2.5-Pro pipeline achieves the strongest event alignment, with 45.7\% abnormality recall and perfect 100\% duration recall and coverage, despite lower SummaC and clarity scores. Overall, these results demonstrate that standard NLP metrics can substantially overestimate clinical reliability, whereas event-based evaluation directly exposes omission and grounding failures in RTM summarization.

\section{Discussion and Conclusion}

Our results reveal three central insights about multimodal clinical time-series summarization.

\textbf{The ``Fluency Illusion'': Standard metrics can mask clinical omission.}
Zero-shot text-only models achieve moderate semantic scores and high clarity ratings, yet exhibit near-zero abnormality recall (0.0--2.1\%) and no duration detection, showing that fluent narratives may still omit critical physiological events. In safety-critical RTM settings, such omission errors are particularly concerning, yet they remain largely invisible to surface-level summarization metrics.

\textbf{Statistical conditioning mitigates, but does not resolve, numeric grounding limitations.}
Providing threshold-aware summary statistics substantially improves factual alignment acorss different models. For example, Llama-3-8B increases from 0.0\% to 33.3\% abnormality recall and achieves an 52.8\% gains in measurement coverage. This suggests that a key failure mode in raw prompting stems from the difficulty of aggregating long sequences of numeric time-series tokens. However, even with structured conditioning, many clinically defined abnormal events remain unreported, indicating that prompt-based scaffolding alone cannot guarantee the completeness required for automated triage.

\textbf{Visualization-grounded reasoning yields the strongest event-level alignment.}
The image-based Gemini-2.5-Pro pipeline achieves the highest abnormality recall (45.7\%), perfect duration recall (100\%), and complete modality coverage (100\%). Interestingly, these gains occur alongside lower conventional semantic scores. This decoupling suggests that visual grounding may promote more faithful detection of sustained physiological deviations, even when linguistic polish is reduced. For multimodal monitoring, perceptual grounding appears to support event sensitivity more effectively than purely textual conditioning.

Together, these findings validate the necessity of event-based evaluation for clinical summarization. By grounding assessment in transparent rule-derived abnormality and duration events, our framework exposes critical reliability gaps that standard similarity-based metrics obscure. Future work in clinical time-series summarization should prioritize \emph{structural event alignment} as a prerequisite for trustworthy deployment in remote therapeutic monitoring.

\bibliography{iclr2026_conference}
\bibliographystyle{iclr2026_conference}

\appendix
\section{Appendix}
\subsection{LLM Usage}
Large language models (LLMs) were used only as general-purpose writing aids to improve the clarity and presentation of the manuscript. In particular, LLMs assisted with grammar, style, and suggested alternative phrasings for prompt instructions. All scientific contributions, including the research ideas, experimental design, and core arguments, were developed by the authors, and all factual claims were independently verified.

\subsection{Dataset Acknowledgement}
This study uses the TIHM (Technology Integrated Health Management) dataset for remote healthcare monitoring in dementia\cite{palermo2023tihm}. We acknowledge the Surrey and Borders Partnership NHS Foundation Trust for providing access to the dataset and for supporting its use in research.
\subsection{Related Works}
\paragraph{Time-series foundation models and remote monitoring.}
Recent time-series foundation models such as Lag-Llama, Chronos, and TimesFM learn generic temporal representations that transfer across forecasting domains~\citep{rasul2023lag,ansari2024chronos,das2024decoder}. In parallel, the TIHM project has demonstrated that dense in-home monitoring for people living with dementia can support early detection of deterioration through multimodal physiological and behavioural signals~\citep{palermo2023tihm}. Our work sits at the intersection: rather than proposing a new forecasting backbone, we treat TIHM-1.5 as a testbed for evaluating how well existing LLMs and vision–language models can turn RTM time series into clinically faithful daily narratives.

\paragraph{Clinical Summarization with Large Language Models.}
Large language models have recently been applied to clinical summarization tasks, including discharge summaries, radiology reports, and longitudinal EHR compression~\citep{van2023clinical,van2024adapted}. While these systems demonstrate promising linguistic fluency and task-specific performance, evaluation typically relies on ROUGE-style overlap metrics or clinician preference studies\cite{fraile2025expert, vasilev2025evaluating}. In contrast, our setting involves multimodal physiological time series rather than free-text EHR inputs, and we focus explicitly on whether generated summaries capture threshold-defined abnormalities and sustained deviations — properties that are not directly measured by conventional summarization benchmarks.

\paragraph{Faithfulness and physiological reasoning.}
A large body of work has highlighted gaps between surface-level summarization quality and factual consistency, leading to metrics such as SummaC and AlignScore that compare generated text against source documents using learned entailment or alignment models~\citep{laban-etal-2022-summac,zha-etal-2023-alignscore}. However, these metrics operate over unstructured text and do not directly assess whether specific physiological events are correctly captured. Our event-based evaluation follows this spirit in the RTM setting: by grounding abnormalities and sustained deviations in transparent rules over time series, we expose omission and hallucination errors that are largely invisible to standard semantic faithfulness scores. A further challenge in remote monitoring is that physiological streams are affected by artifacts, missingness, and device-dependent noise~\citep{shcherbina2017accuracy,bent2020investigating}. Prior work has proposed signal-quality indices for ECG/PPG to quantify reliability in ambulatory monitoring~\citep{orphanidou2014signal}, and has empirically analyzed systematic error sources in wearable optical heart-rate sensing~\citep{bent2020investigating}. Because such imperfections can cause both false alarms and missed abnormalities, our event-based protocol is intentionally conservative: it evaluates whether summaries explicitly capture threshold-defined abnormalities and sustained deviations, while separately measuring modality coverage to expose omissions under realistic sensing conditions~\citep{kompa2021second}. More broadly, high-stakes clinical AI motivates evaluation protocols that are transparent and auditable rather than purely similarity-based. Interpretability arguments in such settings emphasize methods that can be scrutinized and verified~\citep{rudin2019stop,guo2017calibration,ovadia2019can,geifman2019selectivenet}. Our rule-grounded event library and strict matching criteria follow this spirit by making the evaluation target explicit and reproducible~\citep{wachter2017counterfactual}.

\paragraph{Physiological waveform and wearable foundation models.}
Beyond generic time-series forecasting, recent work has trained large self-supervised foundation models directly on physiological biosignals such as ECG and PPG. For example, ECG-FM provides an open foundation model for electrocardiograms, enabling transfer across downstream clinical ECG tasks~\citep{mckeen2025ecg}. Complementarily, large-scale pretraining on consumer wearable biosignals has been demonstrated using hundreds of thousands of participants’ PPG and ECG recordings, showing that representations learned without labels can encode clinically relevant attributes and conditions~\citep{abbaspourazad2023large}. Datasets that pair ECG signals with question-answering supervision (rather than only diagnostic labels) have also been proposed as a step toward more structured reasoning interfaces~\citep{oh2023ecg}. These efforts are closely related to our setting in that they target physiological sensing at scale; however, our focus is not representation learning, but \emph{clinically faithful narrative summarization} and \emph{event-level verifiability} from daily remote-monitoring time series.

\subsection{Prompt Templates}

\begin{figure}[!ht]
\centering
\begin{tcolorbox}[
    title=Prompt: Zero-Shot,
    colframe=black!50,
    colback=black!5,
    boxrule=0.5pt,
    arc=2pt,
    fonttitle=\bfseries,
]
\noindent
You are a clinical AI assistant specializing in remote monitoring for dementia patients. 
Your task is to provide a concise, clinically relevant summary of a patient's raw time-series data for a doctor.

\medskip
\noindent
\textbf{Patient data block:}\\
\texttt{PATIENT DATA FOR \{target\_date.strftime('\%\!Y-\%\!m-\%\!d')\}:}\\
\texttt{---}\\
\texttt{\{structured\_patient\_text\}}\\
\texttt{---}

\medskip
\noindent
\textbf{Instructions:} You MUST adhere strictly to the data provided in the ``PATIENT DATA'' section. 
Your primary goal is factuality. Based only on the raw time-series data provided, generate a summary that addresses the following:

\begin{enumerate}
    \item \textbf{Grounding:} Every statement you make MUST be directly supported by a data point in the provided text. 
    Do not infer trends from single data points. Do not add information that is not present.
    \item \textbf{Handling Missing Data:} If a category like ``Sleep Patterns'' is missing from the input, you MUST state: 
    ``No data available for this category.''
    \item \textbf{Overall Status:} Provide a one-sentence overview of the patient's day.
    \item \textbf{Physiological Analysis:} Analyze the time-series vitals. 
    Note any trends, stability, or significant spikes/dips throughout the day.
    \item \textbf{Behavioral Analysis:} Analyze the activity and sleep logs. 
    Describe the patient's routine (e.g., when they were active, when they slept) and sleep quality.
    \item \textbf{Clinically Significant Events:} If a labeled event is present, you MUST highlight it and 
    try to correlate it with the sensor data.
\end{enumerate}

\noindent
Format the output as a clean, bulleted list. Be specific and refer to times if necessary.

\end{tcolorbox}
\end{figure}

\begin{figure}[!ht]
\centering
\begin{tcolorbox}[
    title=Prompt: Statistical,
    colframe=black!50,
    colback=black!5,
    boxrule=0.5pt,
    arc=2pt,
    fonttitle=\bfseries,
]
\small
\noindent
You are a clinical summarization assistant trained to generate safe, factual, and structured
remote monitoring reports for elderly patients with dementia. You operate under clinical
supervision and your outputs will be evaluated by physicians for factual accuracy,
actionability, and clarity.

\medskip
\noindent
\textbf{Patient data block:}
\begin{flushleft}
\texttt{PATIENT DATA FOR \{target\_date\}:}\\
\texttt{---}\\
\texttt{\{structured\_patient\_text\}}\\
\texttt{---}
\end{flushleft}

\medskip
\noindent
\textbf{Instructions.} You MUST only use and infer information explicitly present in the
\emph{PATIENT DATA} section. You are NOT allowed to fabricate, generalize, or assume any
patterns without specific supporting data. You must flag any uncertainty or data absence
explicitly.

\medskip
\noindent
Format your output using the following structure:

\medskip
\noindent
\textbf{OVERALL STATUS}
\begin{itemize}
    \item One-sentence overview of the patient's day.
\end{itemize}

\medskip
\noindent
\textbf{PHYSIOLOGICAL ANALYSIS}

For each vital sign (Heart Rate, Systolic/Diastolic Blood Pressure, Body Temperature), perform:

\begin{enumerate}
    \item \textbf{Abnormality Check:} Compare average and peak values to the ranges below.
    If outside range, state as ``Abnormally High'' or ``Abnormally Low'' and include specific values:
    \begin{itemize}
        \item Heart Rate: 50--90 bpm
        \item Systolic BP: 90--140 mmHg
        \item Diastolic BP: 60--90 mmHg
        \item Temperature: 35.0--37.5\,$^\circ$C
    \end{itemize}
    \item \textbf{Trend Analysis:} Identify any clear increasing or decreasing trends over several
    hours, supported by multiple timestamps. Flag uncertain or noisy data.
    \item \textbf{Duration Analysis:} If abnormalities were sustained over consecutive readings
    (e.g.\ $>$30 minutes), report duration and time range.
\end{enumerate}

\medskip
\noindent
\textbf{BEHAVIORAL ANALYSIS}
\begin{itemize}
    \item Summarize daily activity patterns and sleep data.
    \item Include periods of peak movement or long inactivity.
    \item For sleep, report total duration and breakdown by sleep stage (if available).
    \item Flag any missing data explicitly (e.g.\ ``No data available for sleep patterns'').
\end{itemize}

\medskip
\noindent
\textbf{CLINICALLY SIGNIFICANT EVENTS}
\begin{itemize}
    \item If labeled events are present (e.g., Agitation, Fall), describe timing and attempt
    correlation with physiology or behavior.
\end{itemize}

\medskip
\noindent
Ensure all bullet points are supported by timestamped data. Do not infer anything not backed
by the provided input.

\end{tcolorbox}
\end{figure}

\begin{figure}[!ht] 
\centering
\begin{tcolorbox}[
    title=Prompt: Vision Based,
    colframe=black!50,
    colback=black!5,
    boxrule=0.5pt,
    arc=2pt,
    fonttitle=\bfseries,
]
\noindent
\textbf{Prompt: Vision-Based Clinical Summarization}

You are a clinical summarization assistant trained to generate safe, factual, and structured remote monitoring reports for elderly patients with dementia. Your outputs are evaluated for factual accuracy, actionability, and clarity.

The provided image contains a single vital sign time series for \{\texttt{signal}\}. You must report only what is explicitly visible in the image.

You MUST use only information directly supported by the visual data. Do not fabricate, generalize, or assume patterns without clear supporting evidence. Explicitly state any uncertainty or missing information.

Your response MUST follow this structure exactly:

\textbf{OVERALL STATUS}  
Provide a one-sentence overview of the patient’s day based only on this \{\texttt{signal}\} plot.

\textbf{PHYSIOLOGICAL ANALYSIS}  
Perform:
\begin{itemize}
    \item Abnormality check relative to clinical thresholds.
    \item Trend analysis across multiple timestamps.
    \item Duration analysis for sustained abnormalities ($>$30 minutes).
\end{itemize}

For abnormality reporting, you MUST include the vital name exactly as written: ``\{\texttt{signal}\}''.  
You MUST use one of the following exact templates:
\begin{itemize}
    \item ``\{\texttt{signal}\} was Abnormally High (value: X.X).''
    \item ``\{\texttt{signal}\} was Abnormally Low (value: X.X).''
    \item ``\{\texttt{signal}\} was within normal range.''
\end{itemize}

Clinical reference ranges:
\begin{itemize}
    \item Heart Rate: 50--90 bpm
    \item Systolic BP: 90--140 mmHg
    \item Diastolic BP: 60--90 mmHg
    \item Temperature: 35.0--37.5~$^\circ$C
\end{itemize}

\textbf{BEHAVIORAL ANALYSIS}  
If activity or sleep context is visible, summarize it. Otherwise state:  
``No data available for activity/sleep from this image.''

\textbf{CLINICALLY SIGNIFICANT EVENTS}  
If labeled events are visible, describe timing and correlation. Otherwise state:  
``No labeled events visible.''

Every statement must be grounded in timestamped data visible in the image. Do not infer anything not supported by the provided visual evidence.

\end{tcolorbox}
\end{figure}
\FloatBarrier
\subsection{Case Study: Omission/misclassification of systolic BP abnormality}
We present a case that illustrates the type of safety-critical failure our evaluation framework is designed to detect: the omission and misclassification of a clinically significant abnormality despite a fluent and confident narrative in the LLM-generated summary. On this patient day, our event-based ground truth identifies an abnormal systolic blood pressure (SBP) of 177 mmHg, which exceeds the predefined threshold of 140 mmHg. However, the LLM summary states that “Blood pressure remained within normal limits,” directly contradicting the physiological evidence.

This error reflects both omission, because the elevated SBP is not explicitly mentioned, and misclassification, because the summary provides a reassuring statement that contradicts the underlying data. In clinical practice, such an inaccurate summary could obscure a clinically relevant signal and potentially delay appropriate patient intervention.

Notably, the LLM-generated summary remains linguistically coherent and topically aligned with the concept of blood pressure, meaning that conventional similarity-based evaluation metrics could still rate the summary as acceptable. By explicitly aligning narrative claims with transparent, threshold-defined clinical events, our event-based evaluation framework makes this failure detectable. This example demonstrates that the omission of true abnormalities—rather than overt hallucination—may represent the primary safety risk in RTM time-series summarization.
\begin{tcolorbox}[
    colframe=black!50,
    colback=black!5,
    boxrule=0.5pt,
    arc=2pt,
    left=6pt,right=6pt,top=6pt,bottom=6pt
]
\captionof{table}{\textbf{Case study: omission/misclassification of systolic BP abnormality.}
Our event-based evaluation flags this day as \emph{Abnormally High} SBP, while the text-only LLM summary incorrectly reports normal blood pressure.}
\label{tab:case_study_sbp_failure}

\small
\textbf{Patient:} 8a835 \hfill
\textbf{Date:} 2019-04-20

\medskip
\textbf{Ground-truth fact (rule-derived):}
\begin{itemize}
    \item \textbf{Systolic BP = 177 mmHg} (\textbf{Abnormally High}; threshold $>140$ mmHg)
\end{itemize}

\medskip
\textbf{LLM summary (excerpt):}
\begin{quote}
\emph{``Blood pressure remained within normal limits.''}
\end{quote}

\medskip
\textbf{Evaluation outcome (ours):}
\begin{itemize}
    \item \textbf{Abnormality recall for SBP:} Missed (no explicit mention of high/elevated SBP)
    \item \textbf{Failure type:} \textbf{Omission / misclassification} (reassuring statement contradicts rule-derived abnormality)
\end{itemize}

\medskip
\textbf{Why this matters:}
\begin{itemize}
    \item Conventional fluency/faithfulness metrics can remain high even when the summary
    \emph{fails to surface a clinically significant event}.
    \item This illustrates the primary reliability gap our evaluation captures: \textbf{event-level omission not penalized by generic text metrics}.
\end{itemize}
\end{tcolorbox}

\subsection{Case Study: Mischaracterization of systolic BP abnormality (wrong magnitude/timing)}
We present a case that illustrates a safety-relevant failure mode our evaluation framework is designed to detect: a clinically meaningful abnormality is mentioned, but its severity and temporal localization are incorrect, creating a misleading clinical picture despite a fluent narrative.

For patient d7a46 on 2019-06-11, the raw physiology contains multiple systolic blood pressure (SBP) readings far above the predefined threshold of 140 mmHg. In particular, SBP reaches 188 mmHg at 14:36 and remains in a markedly elevated range (approximately 173–183 mmHg) across several measurements between 14:36 and 17:55. However, the LLM-generated summary reports a lower peak SBP of 150 mmHg and assigns the abnormal episode to a different time window (13:00–13:30). This is not an omission of the concept of “high BP,” but rather a mischaracterization of the event’s magnitude and timing.

In clinical practice, underestimating the peak SBP and shifting the episode to an incorrect time interval can affect downstream interpretation, including severity assessment, correlation with activity or medication, and escalation decisions. This example motivates event-level evaluation that verifies not only whether an abnormality was mentioned, but also whether the reported values and time intervals accurately match the underlying time-series evidence.

\begin{figure}[!htbp]
    \centering
    \includegraphics[width=0.95\linewidth]{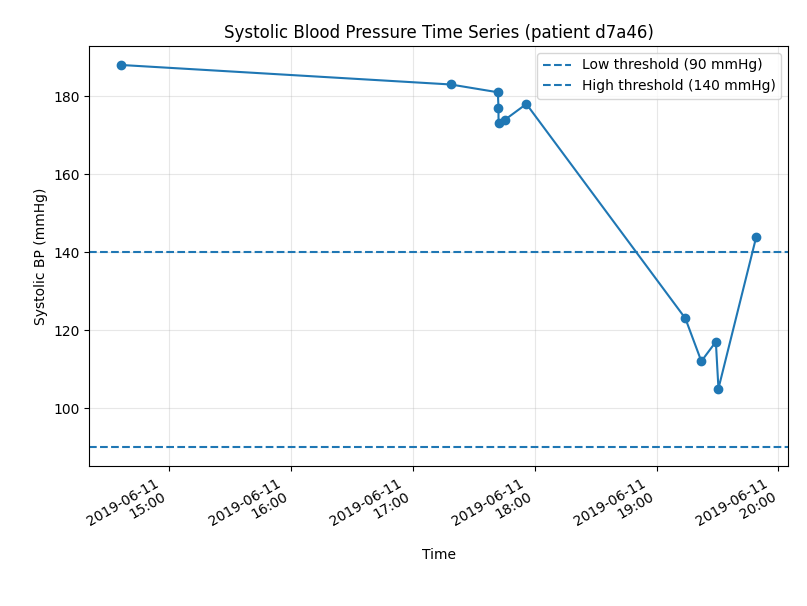}
    \caption{\textbf{Systolic blood pressure (SBP) over time for patient d7a46 on 2019-06-11.}
    The SBP peaks at 188 mmHg (14:36) and remains markedly elevated across multiple readings through 17:55,
    contradicting the summary’s reported peak (150 mmHg) and time window (13:00--13:30).}
    \label{fig:case_sbp_d7a46_2019_06_11}
\end{figure}
\begin{tcolorbox}[
    colframe=black!50,
    colback=black!5,
    boxrule=0.5pt,
    arc=2pt,
    left=6pt,right=6pt,top=6pt,bottom=6pt
]
\captionof{table}{\textbf{Case study: mischaracterization of systolic BP abnormality (magnitude/timing mismatch).}
Our event-based evaluation flags this day as \emph{Abnormally High} SBP based on the raw time-series, while the LLM summary reports a lower peak and an incorrect time window.}
\label{tab:case_study_sbp_mismatch}

\small
\textbf{Patient:} d7a46 \hfill
\textbf{Date:} 2019-06-11 \hfill

\medskip
\textbf{Ground-truth evidence (rule-derived from time-series):}
\begin{itemize}
    \item \textbf{SBP threshold:} Abnormally High if \(>140\) mmHg
    \item \textbf{Observed SBP values (subset):}
    14:36 \(\rightarrow\) \textbf{188},
    17:18 \(\rightarrow\) \textbf{183},
    17:41 \(\rightarrow\) \textbf{181},
    17:42 \(\rightarrow\) \textbf{177/173},
    17:45 \(\rightarrow\) \textbf{174},
    17:55 \(\rightarrow\) \textbf{178} (mmHg)
    \item \textbf{Peak SBP:} \textbf{188 mmHg} at \textbf{14:36}
    \item \textbf{Temporal localization:} markedly elevated readings recur across \textbf{14:36--17:55}
\end{itemize}


\medskip
\textbf{LLM summary (excerpt):}
\begin{quote}
\emph{``The systolic blood pressure was abnormally high at 150 mmHg (peak value) \dots sustained \dots from 13:00 to 13:30.''}
\end{quote}

\medskip
\textbf{Evaluation outcome (ours):}
\begin{itemize}
    \item \textbf{Abnormality detection (SBP):} Partially correct (mentions “abnormally high SBP”)
    \item \textbf{Value accuracy:} \textbf{Incorrect} (reported peak 150 vs observed peak 188)
    \item \textbf{Temporal accuracy:} \textbf{Incorrect} (reported 13:00--13:30 vs observed elevated measurements 14:36--17:55)
    \item \textbf{Failure type:} \textbf{Mischaracterization} (wrong magnitude / wrong time window)
\end{itemize}

\medskip
\textbf{Why this matters:}
\begin{itemize}
    \item A summary can sound clinically plausible while \emph{understating severity} and \emph{misplacing timing},
    which can distort clinician interpretation and downstream decision-making.
    \item This highlights a key reliability gap beyond binary “mentioned vs omitted”:
    \textbf{event fidelity requires correct values and intervals, not just correct topic}.
\end{itemize}
\end{tcolorbox}

\end{document}